\documentclass[sn-basic,iicol]{sn-jnl}
\usepackage{balance}
\usepackage{xcolor}
\usepackage{hyperref}
\definecolor{hc}{RGB}{0,0,255} 
\usepackage{bm}
\usepackage{algorithm}
\usepackage{algpseudocode}
\usepackage{amsmath}

\jyear{2022}
\theoremstyle{thmstyleone}%

\theoremstyle{thmstyletwo}%
\theoremstyle{thmstylethree}%

\begin{document}

\title{SE-GCL: An Event-Based Simple and Effective Graph Contrastive Learning for Text Representation}

\author[1]{\fnm{Tao} \sur{Meng}}\email{mengtao@hnu.edu.cn}

\author[1]{\fnm{Wei} \sur{Ai}}\email{aiwei@hnu.edu.cn}

\author[1]{\fnm{Jianbin} \sur{Li}}\email{jianbinli@csuft.edu.cn}

\author*[1]{\fnm{Ze} \sur{Wang}}\email{fufangze@csuft.edu.cn}

\author[1]{\fnm{Yuntao} \sur{Shou}}\email{shouyuntao@stu.xjtu.edu.cn}

\author[2]{\fnm{Keqin} \sur{Li}}\email{lik@newpaltz.edu}

\affil[1]{\orgdiv{College of Computer and Mathematics}, \orgname{Central South University of Forestry and Technology}, \orgaddress{\city{Hunan}, \state{Changsha} \postcode{410004},  \country{China}}}
\affil[2]{\orgdiv{Department of Computer Science}, \orgname{State University of New York}, \orgaddress{\state{New Paltz}, \city{New York 12561},  \country{USA}}}

\abstract{Text representation learning is significant as the cornerstone of natural language processing. In recent years, graph contrastive learning (GCL) has been widely used in text representation learning due to its ability to represent and capture complex text information in a self-supervised setting. However, current mainstream graph contrastive learning methods often require the incorporation of domain knowledge or cumbersome computations to guide the data augmentation process, which significantly limits the application efficiency and scope of GCL. Additionally, many methods learn text representations only by constructing word-document relationships, which overlooks the rich contextual semantic information in the text. To address these issues and exploit representative textual semantics, we present an event-based, simple, and effective graph contrastive learning (SE-GCL) for text representation. Precisely, we extract event blocks from text and construct internal relation graphs to represent inter-semantic interconnections, which can ensure that the most critical semantic information is preserved. Then, we devise a streamlined, unsupervised graph contrastive learning framework to leverage the complementary nature of the event semantic and structural information for intricate feature data capture. In particular, we introduce the concept of an event skeleton for core representation semantics and simplify the typically complex data augmentation techniques found in existing graph contrastive learning to boost algorithmic efficiency. We employ multiple loss functions to prompt diverse embeddings to converge or diverge within a confined distance in the vector space, ultimately achieving a harmonious equilibrium. We conducted experiments on the proposed SE-GCL on four standard data sets (AG News, 20NG, SougouNews, and THUCNews) to verify its effectiveness in text representation learning. The accuracy achieved on the respective datasets is 91.56\%, 86.76\%, 98.03\%, and 97.79\%, demonstrating superior performance on most datasets compared to baseline methods.}


\maketitle

\vspace{5pt}

\section{Introduction}\label{sec1}
Text representation learning is a fundamental aspect of natural language processing that helps capture semantic and syntactic nuances of textual data. It enables adequate comprehension and generation by machine learning models \cite{xu2023cnn, shou2022conversational, shou2025masked, shou2022object, shou2023comprehensive, shou2024adversarial, meng2024deep, shou2023adversarial, ai2023gcn}. Its paramount importance lies in its transformative capacity to bridge the gap between raw text data and computational models, paving the way for advancements in information retrieval, text mining, machine translation, sentiment analysis, and automated reasoning \cite{li2022survey, ai2024gcn, meng2024multi, shou2024contrastive, shou2024spegcl, shou2024efficient, ying2021prediction}. The pervasive nature of textual data in the digital age underscores the necessity for advanced text representation learning methods. Despite their significant progress, existing text representation methods grapple with several challenges.

\begin{figure}
	\centering
	\includegraphics[width=0.48 \textwidth]{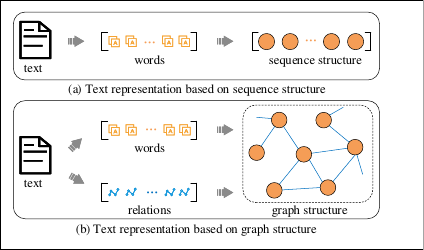}
	\caption{Different processing methods for text representation.}
	\label{Fig1}
\end{figure}

One major flaw of the prevailing approach is its tendency to treat text as an undifferentiated sequence and extract only keywords or sentences as representatives of the entire text. At best, it connects other similar texts or additional information to enhance data. These approaches significantly oversimplify the inherent content complexity and discount its contextual richness. Traditional word-based representations such as BoW and TF-IDF effectively ignore the order of words, while others like Word2Vec \cite{mikolov2013efficient}, GloVe \cite{pennington2014glove}, and fastText \cite{joulin2017bag} model the semantics of individual words but struggle with capturing the nuances of longer phrases or sentences. Numerous academics are diligently exploring sentence-level representations. For instance, MixCSE \cite{zhang2022unsupervised} forces the model to capture subtle sentence semantic features by introducing hard negative examples. However, the encapsulation of text that encompasses multiple sentences invariably results in the erosion of structural integrity and the dilution of long-range semantic coherence. Even more advanced Transformer-based \cite{vaswani2017attention, shou2023graph, shou2023czl, meng2024masked, shou2024revisiting, ai2024edge, zhang2024multi} methods treat the entire text as a sequence of words and then employ the attention mechanism to understand the context. For example, Bert \cite{devlin2018bert}, pre-trained on a large-scale corpus, uses the multi-head attention mechanism to capture the dependencies between words and achieve competitive results in multiple natural language tasks. ELECTRA \cite{Clark2020ELECTRA} designs the replaced token detection method to achieve better robustness with lower training costs. However, Transformer-based models mainly focus on token information and may ignore the complex interconnections and multiple-level hierarchy. In addition, it has a length limit on the input text, and truncation of the text may lead to unpredictable loss of text semantics. Its processing method is shown in Fig. \ref{Fig1}(a). While powerful, such models may fail to consider the high-level semantic structure inside the document, limiting their effectiveness in text representation learning.

The advent of graph neural networks \cite{welling2016semi, ai2023two, meng2024revisiting, ai2024mcsff, shou2023graphunet, ai2024graph, shou2024low, shou2024graph, ai2024seg, ai2024contrastive, fu2024sdr, shou2024dynamic} provides a new perspective for text representation, making it possible to model unstructured data such as text. Its processing method is shown in Fig. \ref{Fig1}(b). For example, TextGCN \cite{yao2019graph} builds a corpus-level heterogeneous graph and uses word nodes as a bridge for message passing to learn the representation of document nodes. Although it solves the problem of converting a text corpus into a graph, it cannot take advantage of the rich contextual information in the text. TextING \cite{zhang2020every} builds a separate graph for text, reducing memory consumption, but it ignores the rich relationship information between entities in the text and lacks the grasp of semantic information. In summary, both sequence-based and graph-based methods do not fully utilize entities and their relationship information, which cannot represent semantically sparse text well. Therefore, designing a text representation method that can truly embrace natural language's semantic and structural complexity is still a problem worth exploring.

Unsupervised graph contrastive learning has been applied in text representation learning in the ongoing pursuit of a more streamlined learning paradigm. The advantage of GCL-based methods is that they can autonomously identify underlying structural features in data without annotating the data. For example, CGA2TC \cite{yang2022contrastive} constructs a corpus-level graph with words and documents as nodes and designs a contrastive graph representation framework with an adaptive augmentation strategy, which can effectively remove graph noise and achieve promising performance. However, it does not consider the structural and semantic information in the text comprehensively, leading to less distinguishable representations of texts. To alleviate this problem, TGNCL \cite{li2023graph} constructs a word graph for each text, which captures the rich contextual information of the text. Then, a contrastive learning regularization is developed based on the constructed text graphs to improve the robustness of text representation. Moreover, the efficiency of CGA2TC \cite{yang2022contrastive} and TGNCL \cite{li2023graph} is notably reduced by their reliance on intricate graph data augmentation techniques, including the creation and encoding of contrastive views. Therefore, the delicate equilibrium between robustness and efficiency underscores the urgent need to develop more refined yet effective data augmentation techniques in graph contrastive learning to unlock its latent potential in text representation.

In response to these challenges, we present an event-based, simple, yet effective graph contrastive learning framework SE-GCL for text representation learning. This method diverges from traditional techniques by focusing on textual events as the primary unit of analysis rather than merely extracting keywords and sentences. SE-GCL captures the core intent of texts semantically and structurally by defining textual events and building internal relational graphs for each text. Further, we introduce a streamlined, unsupervised graph contrastive learning framework to leverage the complementarity between semantic and structural textual information for comprehensive feature extraction. Specifically, to improve text representation efficiency, we first mine the event skeletons in the internal relationship graph to preserve only the more essential semantics. Furthermore, we propose a simplification of the complex data augmentation process commonly found in existing graph contrastive learning. For anchor embeddings, instead of GCN, we use MLP to generate anchor embeddings infused with semantic information. For event skeletons, we adopt GCN for embedding representation. This approach explores the complementarity of semantic and structural information while effectively simplifying the strategy for generating embeddings. In another simplification step, we shuffle the anchor embeddings to generate negative embeddings, avoiding the need for more computationally expensive strategies. Lastly, we can achieve equilibrium by manipulating various embeddings using multiple loss functions to approach or diverge from each other within a finite distance in the vector space. This systematic yet innovative approach effectively addresses the challenges current text representation methods face, offering a more efficient and robust path forward. Our source code is available at \href{https://github.com/KrisWongz/SEGCL}{https://github.com/KrisWongz/SEGCL}. The main contributions of this paper can be summarized as follows:

\begin{itemize}
	\item
	We first propose the definition of textual events and construct an event-based internal relational graph to express the core intent of each text at both semantic and structural levels.
	\item
	We propose an event-based graph contrastive text representation learning framework, which can explore the complementarity between semantic and structural information to obtain semantic-rich text representation and achieve better efficiency.
	\item
	Experiments and analysis on real-world datasets show that our method outperforms existing methods in effectiveness and interpretation.
\end{itemize}

\section{Related Work}
\subsection{Word-sentence-based Text Representation}
Primarily, the encoding of textual information hinges on the representation of words, a foundational pillar in the landscape of natural language processing that maintains its indelible significance. Conventional word-based representations like Bag of Words and Term Frequency-Inverse Document Frequency tend to overlook the sequential arrangement of words. In contrast, alternative techniques such as Word2Vec, GloVe, and fastText \cite{mikolov2013efficient, pennington2014glove, joulin2017bag} endeavour to encapsulate the semantics of individual words yet grapple with capturing the subtle intricacies of elongated phrases or sentences. These methodologies have recorded noteworthy successes in the realm of word representation. However, their direct application for text representation poses a formidable challenge.

An advanced approach lies in the representation of sentences, a technique poised to assimilate more robust features. For instance, Zhang \emph{et al.} \cite{zhang2022unsupervised} have proposed a contrastive model, which expands upon SimCSE \cite{gao2021simcse} by iteratively crafting hard negatives through a blend of both positive and negative features. Similarly, Yan \emph{et al.} \cite{yan2021consert} have introduced a Contrastive Framework for Self-Supervised Sentence Representation Transfer, employing contrastive learning to refine BERT \cite{devlin2018bert} in an unsupervised yet efficacious manner. In another noteworthy research \cite{zheng2021sentence}, a multi-layer semantic representation network is explicitly devised for sentence representation, wherein a multi-attention mechanism garners the semantic information across varying sentence levels. Sentence representation has seen substantial advancements and has been effectively incorporated across multiple domains in recent years. Nevertheless, text representation spanning multiple sentences invariably invites the degradation of structural integrity and a concurrent dilution of long-range semantic coherence.

Word-based and sentence-based models achieve superior results in short text sequence representation learning. However, when dealing with lengthy texts that have complex meanings, these models often struggle to grasp the deeper semantic features. This is because they tend to analyze words and sentences in isolation without considering how these elements interrelate or how they contribute to the overall coherence of the text.

\subsection{Text Representation via Deep Learning}
The field of text representation via deep learning methodologies has undergone a remarkable metamorphosis, marked by an exponential surge in model intricacy and the multifaceted representations they facilitate. Primitive undertakings centered predominantly on convolutional neural networks (CNNs) \cite{NIPS2012_c399862d} and recurrent neural networks (RNNs) \cite{zaremba2014recurrent}, along with their long short-term memory (LSTM) \cite{shi2015convolutional} offshoot. These frameworks, exploiting the inherently sequential characteristic of textual data, marked a considerable stride forward from their preceding non-contextual counterparts. Regrettably, the CNNs' focus remains tethered predominantly to local information,  overlooking long-range semantic relationships. Concurrently, RNNs and their ilk possess the capacity to consider the sequence in its entirety but display diminishing effectiveness as the sequence length swells. Ultimately, none of these models demonstrate an efficacious capability in abstracting global semantics.

With the inception of attention mechanisms and transformer architectures, the domain of text representation underwent a significant paradigm shift. Transformer-centric models, such as BERT \cite{devlin2018bert} and GPT \cite{radford2018improving}, seized the potential of the attention mechanism to capture dependencies without regard to their proximity within the textual continuum, effectively circumventing the constraints of RNNs and LSTMs. For example, SWCC \cite{gao2022improving} utilizes document-level co-occurrence information of events to learn event representations without additional annotations.

Simultaneously, the rise of graph neural networks (GNNs) \cite{welling2016semi, velivckovic2017graph, chen20v} signaled a promising development in text representation. Uniquely endowed to grasp the structural nuances innate to text, GNNs address a critical gap often neglected by sequence-oriented models.  For instance, TextGCN \cite{yao2019graph} erects a text graph predicated on word co-occurrence and document-word correlations, subsequently employing a Graph Convolutional Network to learn representations. TREND \cite{wen2022trend} proposed the concepts of events and dynamic nodes, which capture the individual and collective characteristics of events, respectively. TextFCG \cite{wang2023text} builds a single graph for all words in each text, labels edges by fusing various contextual relations, and uses GNN and GRU for text classification.

Although these deep learning-based methods are practical and widely used, they all face difficult problems. First, sequence-based models focus on local dependencies of text but cannot fully capture long-term dependencies. Second, although the graph-based models can construct the global structure of the text corpus, they ignore the rich entity information and relationship information within the texts.

\subsection{Contrastive Representation Learning}
Contrastive representation learning represents another significant frontier in developing of advanced text representation techniques. This branch of learning operates on the principle of learning representations by contrasting positive pairs (similar or related instances) against negative pairs (dissimilar or unrelated instances). Such learning mechanisms have shown remarkable success across various applications, including computer vision and natural language processing\cite{xie2022self}.

SimCLR \cite{chen2020simple} extended InfoMax principles to multiple views and maximized Mutual Information (MI) by augmenting the resulting views with data. The InfoGCL \cite{xu2021infogcl} framework reduced mutual information between contrasting parts through the Information Bottleneck principle while maintaining the integrity of task-relevant information at the level of individual modules and the entire framework. Mo \emph{et al.} \cite{mo2022simple} proposed a simple unsupervised graph representation learning method SUGRL, whose multiple loss explored the complementary information between structural and neighbor information to produce more minor generalization errors. GCNSS \cite{miao2022negative} effectively alleviates the negative sampling bias problem in graph contrastive learning by utilizing label information. NCLA \cite{shen2023neighbor} proposed a learnable graph augmentation strategy to produce safer contrasting views. For the field of text representation, CGA2TC \cite{yang2022contrastive} designed an adaptive data enhancement strategy to effectively filter graph noise information. TextGCL \cite{zhao2023textgcl} uses contrastive learning loss to simultaneously train GCN and Bert to learn a more robust text representation. TGNCL \cite{li2023graph} introduces contrastive learning regularization on text-level graphs to learn robust word representations.

Existing graph data augmentation methods can lead to two potential issues. Firstly, graph data augmentation typically involves view generation and view encoding, which incurs significant computational costs. Secondly, modifying graph information in a random manner (such as node dropping and edge dropping) may result in unpredictable semantic loss. Consequently, there is a pressing need to develop a more efficient strategy for data augmentation in this context.

\begin{figure*}
	\centering
	\includegraphics[width=0.99 \textwidth]{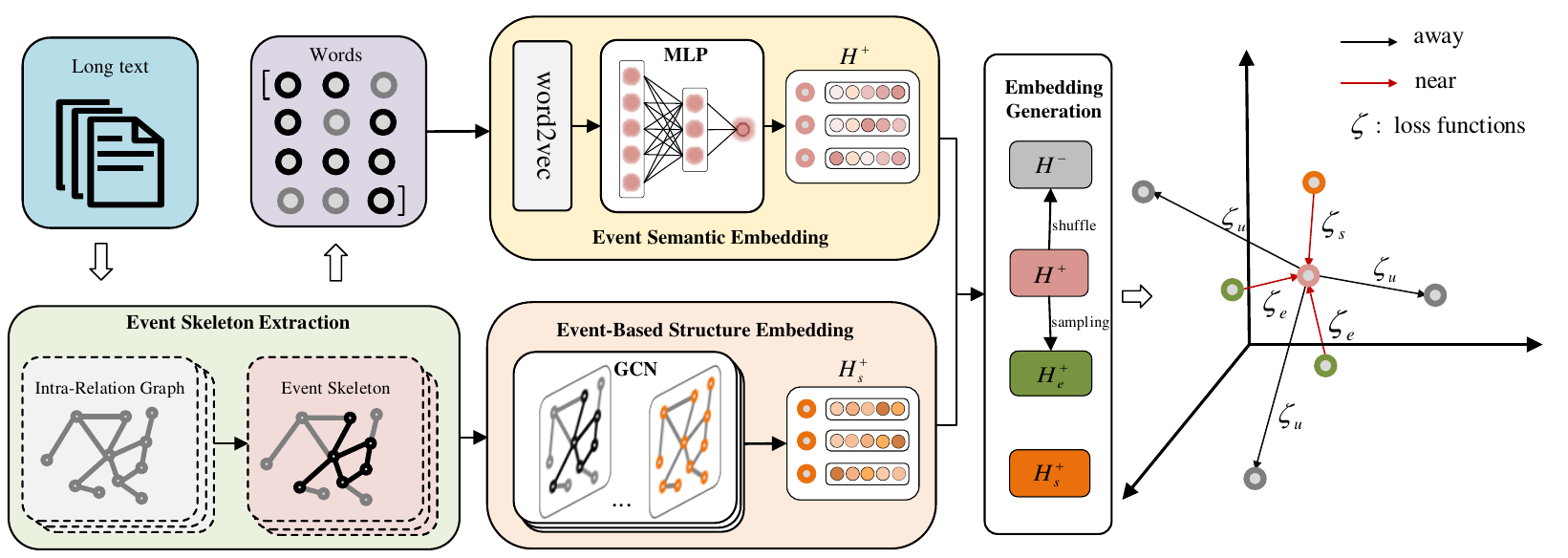}
	\caption{Illustration of our method SE-GCL. First, SE-GCL builds intra-relation graphs for texts and identifies their event skeletons from them. An MLP is used to generate the word node anchor embedding $H^+$ of intra-relation graphs, and the negative embedding $H^-$ is obtained by perturbing the anchor embedding, and the event embedding $H^+_e$ is obtained by sampling. In addition, a GCN will be used to generate structural embeddings $H^+_s$. A contrastive loss is then applied to close the distance between positive embeddings and anchor embeddings while widening the distance between negative embeddings and anchor embeddings.}
	\label{Fig3}
\end{figure*}

\section{Proposed Method}
In this paper, we introduce an event-based, simple yet effective graph contrastive learning (SE-GCL) framework, a novel approach to text representation that effectively captures both the semantic and structural intricacies inherent in natural language. The proposed SE-GCL method comprises four major steps, forming a systematic pipeline for comprehensive text representation, whose overall structure is shown in Fig. \ref{Fig3}.

The first step is the construction of an intra-relation graph. Recognizing that textual events represent core semantic and structural information, we extract these event blocks and build a graph based on their semantic relationships, thereby retaining the most critical and central semantic information of the text.

Next, we introduce the concept of event skeleton extraction. By defining event skeletons and applying them to the intra-relation graphs, we effectively compress and augment them, leading to a more efficient and enhanced representation of textual events.

The third step involves the generation of embeddings in the contrastive framework. We design a streamlined, unsupervised graph contrastive learning framework to exploit the complementarity between semantic and structural textual information for comprehensive feature extraction. We use less complex embedding generation strategies instead of complex data augmentation strategies common in existing graph contrastive learning.

Finally, the SE-GCL method employs multiple loss functions to facilitate the convergence of our model. A harmonious balance is achieved by manipulating various embeddings to approach or diverge from each other within a finite distance in the vector space, ensuring the robustness and effectiveness of our model.

In the following sections, we delve into a detailed exposition of each step, elucidating the innovative mechanisms and strategies that underpin the SE-GCL method.

\subsection{Intra-relation Graph Construction}
\begin{figure}[H]
	\centering
	\includegraphics[width=0.48 \textwidth]{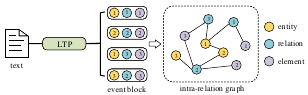}
	\caption{The overall process of constructing intra-relation graph. }
	\label{Fig2}
\end{figure}

In the first stage of our SE-GCL method, we convert the raw text into an intra-relation graph using a syntactic dependency-based Language Technology Platform (LTP) \cite{che-etal-2021-n} event extraction tool. The overall process is shown in Fig. \ref{Fig2}. This tool allows us to delve beyond surface-level syntactic structures of sentences and directly extract deep semantic information, thereby providing a more comprehensive and enriched understanding of the text.

We commence by processing the text into multiple triplet event blocks using the LTP tool. An example of such an event block is [``Peter'', ``eats'', ``apple'']. Each element in these event blocks is referred to as an ``event element''. One of the key advantages of this approach is that we can describe the semantics of sentences through the semantic framework borne by the vocabulary without needing to abstract the vocabulary itself. This is crucial as the number of arguments is invariably smaller than the vocabulary.

Subsequent to the event block formation, we retain the part-of-speech information for each word. This stage can be likened to the process of named entity recognition, wherein entity information, such as person names, place names, and institution names, is identified. The retention of part-of-speech information is crucial as it provides additional context and semantic information that aids in the construction of the intra-relation graph. It allows us to differentiate between entities and actions and to understand the roles different words play within the event blocks.

Building upon the event blocks and the part-of-speech information, we then proceed to construct an intra-relation graph. Initially, we include event blocks with entities, establishing edge connections based on the relationships between these entities. Specifically, entities that appear simultaneously in the same sentence are connected through edges, such as entity 1 and entity 2 in Fig. \ref{Fig2}. It is worth pointing out that the same entity in different event blocks will be treated as a node, which means that different event elements can be connected through the shared event entity.

In the next step, we preserve event blocks where an event element appears multiple times and establish connections based on the co-occurrence relationship of these event elements. It should be noted that the event elements need not be identical for a link to be established. A link is created if the similarity between event elements exceeds a predefined threshold (y). The similarity between event elements is measured using a semantic similarity metric, which takes into account both the semantic and syntactic similarities between the elements. The threshold is determined empirically, with a higher threshold leading to fewer but more confident connections and a lower threshold leading to more but potentially less confident connections.

Through the aforementioned process, we successfully convert the text into an intra-relation graph, capturing the intricate semantic and structural details inherent in natural language. This graph forms the foundation for the subsequent stages of our SE-GCL method.

\subsection{Event Skeleton Extraction}
The second stage of our SE-GCL method involves the extraction of event skeletons. Event skeletons provide a specific event composition architecture frequently observed in these types of articles. The main purpose of event skeleton extraction is to capture representative semantic information within the text by establishing a relation graph between event entities, which helps to mine the core event organization pattern of the text. Therefore, the event skeleton provides an effective way to represent the event structure of text and is an important method to explore rich text contextual semantics.

To extract these event skeletons, we employ the gSpan (graph-based Substructure pattern mining) algorithm \cite{yan2002gspan}, a seminal technique in the field of frequent subgraph mining. The gSpan algorithm operates by mapping graph data to a canonical order string and systematically exploring the search space using a depth-first search strategy. This technique enables us to efficiently identify frequent subgraphs, i.e., substructures that recur at a frequency above a given threshold.

Given a graph $G$, denote the intra-relation graph. In a more sophisticated and elegant manner, we assign identifiers to the nodes and connections within the intra-relation graph. The trio of node categories can be associated with a total of sextet edge types. Subsequently, we arrange the nodes and connections in accordance with the frequency of their identifiers and eliminate those nodes and connections that exhibit a lower frequency, thereby deriving a novel graph, denoted as $G_{new}$. The amalgamation of connections exhibiting a higher frequency into a set, denoted as $E$, can be perceived as the formation of a set $E$ encompassing all connections within the graph $G_{new}$. Arrange the edges in $E$ in descending order of the minimum depth-first search (DFS) encoding order and frequency. Our objective is to discern the subgraph of the frequency within the intra-relation graph, and the connection can also be perceived as a unique subgraph. The connections within the set $E$ can be viewed as the most rudimentary frequent subgraph. The ensuing frequent subgraph mining is predicated upon these frequent connections for recursive mining.

Subsequently, we procure the initial frequent subgraph, denoted as $A$, predicated on $E$, and proceed to augment it recursively. The augmentation process is partitioned into a triad of steps. Initially, an assessment is made to determine whether DFS encoding is fulfilled. If this condition is met, an expansion is executed on the rightmost side. An evaluation of the newly augmented subgraph is conducted to ascertain whether it complies with the support degree. If it does, the recursive expansion continues predicated on the new subgraph. Ultimately, we succeed in obtaining frequent subgraphs.

Once these frequent subgraphs are extracted, they can be represented as event skeletons when applied to the intra-relation graph with event information. As such, the event skeletons encapsulate the representative structure of the events, providing a compact yet comprehensive snapshot of the most salient semantic elements within the text. This efficient representation paves the way for the subsequent stages of the SE-GCL method.

\subsection{Embedding Generation}
We focus on the generation of embeddings within the graph contrastive learning framework in the third phase, which includes the creation of anchor embeddings, positive embeddings, and negative embeddings.

\textbf{Anchor Embedding:} Traditionally, embedding generation methods have relied heavily on deep learning techniques such as GCN, followed by the application of a readout function to obtain anchor embeddings. While effective, these methods can be computationally intensive and time-consuming, posing challenges for scalability.

In contrast, our approach in the SE-GCL method is to leverage the intra-relation graph directly to generate anchor embeddings with event information. Specifically, we employ a simple  multi-layer perceptron (MLP) to transform the word nodes in the intra-relation graph into anchor embeddings. This approach effectively reduces the computational burden of the algorithm, thereby enhancing its scalability. By using the MLP, we can capture the event information inherent in the word nodes of the intra-relation graph. The formula is expressed as:
\begin{equation}
	H^{l+1} = sigmod(\sum_{i=0}^M H_i^l \cdot W_i^l + b),
\end{equation}where $H_i^l$ is the embedding of the $l$-th layer and $H^0$ is the input. $ W_i^l$ is the weight of the $l$-th layer. $M$ represents the number of neurons. $b$ is the artificially set bias. Here, we regard the output $H$ of the last layer as the anchor embedding.

\textbf{Negative Embedding:} In contrastive learning, the combination of negative samples containing significantly different features and anchor samples can promote the model to learn highly discriminative representations. In the context of generating negative embeddings, many previous methodologies have relied on intricate negative sampling strategies or have utilized GCN to obtain embeddings after distorting the original graph. While these methods can be effective, they are often complex and time-consuming, posing challenges for scalability and efficiency. In contrast, we adopt a simpler strategy: shuffle the anchor embeddings to generate the negative embeddings. This approach significantly reduces the computational burden of the algorithm. Through this strategy, our method destroys the original order of anchor embeddings, creating a set of negative embeddings with distinct characteristics from the anchor embeddings. This aligns with the objective of contrastive learning, which aims to minimize the similarity between negative pairs (i.e., the anchor and negative embeddings). Furthermore, this simple strategy of generating negative embeddings provides discriminative negative samples while significantly reducing the computational cost by removing the graph neural network. The negative embedding $H^-$ is defined as shown in Equation 2.
\begin{equation}
	H^- = shuffle(H).
\end{equation}

\textbf{Positive Embedding:} The generation of positive embeddings is a crucial aspect of the SE-GCL method. While there are diverse approaches to this task, many existing methods rely on GCN to extract the structural information of the graph or to perform further data augmentation. However, we propose two distinct strategies for generating positive embeddings, aiming to capture more complementary information. These strategies focus on two key types of information: structural information $H^+_s$ and event information $H^+_e$.

Structural Information: The first strategy is designed to capture the structural information inherent in the intra-relation graph. This involves leveraging the relationships between the nodes in the graph, as represented by the edges and their properties. By focusing on this structural information, we can capture the underlying architecture of the events in the text, which is crucial for understanding their context and semantics.

Event Information: The second strategy is focused on capturing the event information represented in the intra-relation graph. This involves leveraging the event blocks and event skeletons that we have extracted in the previous steps. By focusing on the event skeleton information, we can capture the specific compositions of events in the text, which are crucial for understanding the core semantics.

For each intra-relation graph $G$, we introduce a two-layer GCN as an encoder to get the structure embedding. Formally, Let $ A $ represents the adjacency matrix of $ G $ and $ D $ is the degree matrix, where $ D_{ii}=\sum_j {A_{ij}} $. Moreover, each node is connected to itself. Then, the neighbor information is aggregated into $ N_v $ to update the embedding of node $ v $ recursively by the aggregation function $ AGG $. The steps can be expressed as follows:
\begin{equation}
		H_i^{l+1}=\sigma(\widetilde{A} \cdot W^l\cdot \rho \cdot concat(H_i^l, AGG(H_j^l,v_j\in N_{v_i})),
\end{equation}where $\sigma$ represents the activation function, such as Leaky ReLU. $ \widetilde{A} = D^{-\frac{1}{2}} A D^{-\frac{1}{2}} $ is the symmetric normalized adjacency matrix. $ W^l $ is the trainable transformation matrix of the layer $ l $. $\rho$ is the event skeleton's weight.

We select nodes in the event skeleton to sample and obtain their average to obtain the positive embedding with the event information. The sampled positive event embeddings provide a new perspective to encourage the model to learn discriminative representations. The formula is expressed as follows:
\begin{equation}
	H_e^ +  = \frac{1}{k}\sum\limits_{}^{} {({H_i}\lvert {{v_i} \in event\_skeleton} )} ,
\end{equation}
where $k$ represents the number of nodes in the event skeleton.

In general, we design two positive embeddings from the structural and event levels to explore their complementarity. The structure embedding contains the information of the whole intra-relation graph, and the event embedding contains part of the nodes' information. They interpret the same graph from different perspectives. Therefore, it is considered separately, and we can obtain specific complementary information.

\subsection{Multi-Loss Functions}
The purpose of contrastive learning is to make positive embedding close to anchor embedding and negative embedding away from anchor embedding. Since a small generalization error may improve the generalization ability of contrastive learning, and reducing the intra-class variation or expanding the inter-class variation is an effective solution to reduce the generalization error, we design multi-loss functions for the resulting positive, anchor, and negative embeddings based on the event skeleton and intra-relation graph. Meanwhile, we introduce an upper bound loss to improve efficiency and replace the discriminator method. The multi-loss can be formulated as follows:
\begin{equation}
	sim(H, H^+) < sim(H, H^-) - \eta,
\end{equation} where $ sim(\cdot) $ is a similarity measure function, such as $l2$-norm distance, and $ \eta $ is a non-negative number to ensure that the distance between positive and negative embeddings is within a fixed range. Integrate all negative embeddings to get the loss $\zeta$:
\begin{equation}
	\zeta_{multi} = \frac{1}{k} \sum_{i=1}^k \{ sim(H, H^+)^2 - (sim(H, H_i^-)^2 - \eta) \}^{max},
\end{equation}where $ \{\cdot\}^{max} $ means taking the maximum value between $ \{\cdot, 0\} $, k represents the number of negative embeddings.

We apply the loss to the two defined positive embeddings. The loss of structure embedding $H^+_s$ can be formulated as follows:
\begin{equation}
	\zeta_s = \frac{1}{k} \sum_{i=1}^k \{ sim(H, H^+_s)^2 - (sim(H, H_i^-)^2 - \eta) \}^{max},
\end{equation} while the loss of event embedding can be formulated as follows:
\begin{equation}
	\zeta_e = \frac{1}{k} \sum_{j=1}^k \{ sim(H, H^+_e)^2 - (sim(H, H_j^-)^2 - \eta) \}^{max}.
\end{equation}

The implementation of such a multiplet loss, which is essentially a pair of triplet losses, can enhance the disparity between classes upon examining Eqs. (7) and (8), two potential scenarios emerge. The first scenario is one where the loss incurred in Eq. (8) is null, while that of Eq. (7) is non-zero. In this case, Eq. (7) continues to extend the vector representation of the negative sample in comparison to the positive sample of Eq. (7) further afield. The converse scenario is equally plausible. Eq. (8) also serves to distance the vector representation of negative samples. Collectively, these two equations contribute significantly to the differentiation between classes. In this way, we can effectively expand the inter-class variation in the case of one type of loss with poor effect to obtain complementary information of event information and structural information.

In order to avoid the situation where the gap between the anchor embedding and the positive embedding itself is very large, we set an upper bound $\theta$ for the negative pair to ensure that the distance between the negative embedding and the anchor embedding is limited, which can effectively reduce the intra-class variation. The upper bound loss is defined as follows:

\begin{equation}
	\zeta_u = \frac{1}{k} \sum_{i=1}^k \{ sim(H, H^+)^2 - (sim(H, H_i^-)^2 - \eta - \theta) \}^{min},
\end{equation}where $ \{\cdot\}^{min} $ means taking the minimum value between $ \{\cdot, 0\} $. The final multi-loss $\zeta$ is given by

\begin{equation}
	\zeta = W_e\cdot\zeta_e + W_s\cdot\zeta_s + \zeta_u,
\end{equation}where $ W_e $ and $ W_s $ are the weights of $ \zeta_e $ and $ \zeta_w $, respectively. We average all final node embeddings in the graph to get the text representation.

Overall, we explore the complementary information between structural information and neighbor information via two triplet loss functions to amplify inter-class variation and an upper bound loss to reduce intra-class variation. The final multi-loss effectively maximizes the difference between classes and minimizes the difference within classes. This kind of constraint in two directions can reduce the generalization error. The process of SE-GCL can be shown as $Algorithm$ $1$.

\section{Experiments}
We conduct experiments on common datasets to evaluate the performance of the SE-GCL method. In this section, we will introduce the data sets and preprocessing, comparison of methods, experiment settings, results, and corresponding analysis.

\subsection{Data Sets}
In our experiments, we utilized four diverse datasets to evaluate the performance and robustness of the SE-GCL method. These datasets are:

\textbf{AG News}\footnote{\href{http://groups.di.unipi.it/~gulli/AG_corpus_of_news_articles.html}{http://groups.di.unipi.it/~gulli/AG_corpus_of_news_articles.html}}: This is a dataset of news articles from the AG's corpus of news articles on the web, pertaining to the four largest classes. The dataset contains 30,000 training examples and 1,900 test examples per class.

\textbf{20NG}\footnote{\href{http://qwone.com/~jason/20Newsgroups/}{http://qwone.com/~jason/20Newsgroups/}}: This dataset is a collection of approximately 20,000 newsgroup documents partitioned across 20 different newsgroups. It is a popular dataset for experiments in text applications of machine learning techniques, such as text classification and text clustering.

\textbf{SougouNews}\footnote{\href{https://huggingface.co/datasets/sogou_news}{https://huggingface.co/datasets/sogou_news}}: The data of SogouNews is compiled by Sogou Lab. It comes from a total of 1,245,835 news reports from 18 Sohu News channels, including domestic, international, sports, social, and entertainment, from June to July 2012. Considering the device factor and balancing the dataset, we randomly sample 3000 entries in each of the ten categories.

\textbf{THUCNews}\footnote{\href{http://thuctc.thunlp.org/}{http://thuctc.thunlp.org/}}: The ThuCNews corpus is a news document generated by filtering the historical data of the Sina News RSS subscription channel from 2005 to 2011, which contains 14 news categories and about 830,000 news texts. Considering the device factor and balancing the dataset, we randomly sample 5000 entries in each of the 14 categories.

\begin{algorithm}
	\caption{SE-GCL: An Event-Based Simple and Effective Graph Contrastive Learning for Text Representation}
	\label{alg:Framwork}
	\begin{algorithmic}[1]
		\Require
			\bm{$C$}: a text corpus; \bm{$y$}: similarity threshold between event elements.
						
		\Ensure
			\bm{$text\;representation$}: well-encoded embedding.
			
		\State \bm{$eventBlocksSet$} \bm{$\leftarrow$} LTP(\bm{$C$});
		\State \bm{$G$} \bm{$\leftarrow$} BuildGraph(\bm{$eventBlocksSet$}, \bm{$y$});
		\State \bm{$G_{new}$} \bm{$\leftarrow$} gSpan(\bm{$G$});
		\For{$epoch = 1 \to max\_epochs$}
			\State \bm{$H$} \bm{$\leftarrow$} MLP(\bm{$G$});
			\State \bm{$H^-$} \bm{$\leftarrow$} \textit{shuffle}(\bm{$H$});
			\State \bm{$H^+_s$} \bm{$\leftarrow$} GCN(\bm{$G$});
			\State \bm{$H^+_e$} \bm{$\leftarrow$} \textit{sample}(\bm{$H$}, \bm{$G_{new}$});
			\State \bm{$\zeta_s$}, \bm{$\zeta_e$}, \bm{$\zeta_u$} \bm{$\leftarrow$} get loss based on Eq.(7), Eq.(8), Eq.(9) and (\bm{$H$}, \bm{$H^-$}, \bm{$H^+_s$}, \bm{$H^+_e$});
			\State \bm{$\zeta$} \bm{$\leftarrow$} get final multi-loss based on Eq.(10) and (\bm{$\zeta_s$}, \bm{$\zeta_e$}, \bm{$\zeta_u$});
			\State By applying stochastic gradient ascent to update the parameters to minimize \bm{$\zeta$};
		\EndFor
		\State \textbf{Return:} \bm{$text\;representation$}
	\end{algorithmic}
\end{algorithm}

These datasets were chosen due to their diversity in terms of domain (news articles from various categories) and size. This diversity allows us to thoroughly evaluate the performance of the SE-GCL method under different conditions and settings.

\textbf{Preprocessing}: Following previous works, we remove stopwords and low-frequency words (word frequency less than 5), as well as word segmentation operations on all datasets. Apart from this, to manage computational demands and maintain the feasibility of our study, a sampling strategy is applied to the large-scale Chinese datasets.

\subsection{Comparison of Methods}
To evaluate the performance of our method, we compare it with different types of text representation learning methods. Covering different types of models ensures that the evaluation is not biased by specific model types, providing more balanced and representative evaluation results. These baseline methods can be divided into three groups, including Word Embedding Based Models, Sequence Deep Learning Models, and Graph Based Representation Learning Models. The selected methods are as follows:

\noindent \textbf{(1) Word Embedding Based Models}

\textbf{TF-IDF+LR} \cite{yao2019graph}: Bag-of-words model with word frequency inverse document frequency weighting. Use logistic regression as the classifier.

\textbf{fastText} \cite{joulin2017bag}: A simple yet efficient method for text classification (Joulin et al. 2017) that treats the average of word/n-gram embeddings as document embeddings and then feeds the document embeddings into a linear classifier.

\noindent \textbf{(2) Traditional Deep Learning Models}

\textbf{CNN} \cite{NIPS2012_c399862d}: CNN is a type of traditional deep learning model that is commonly used for text classification tasks. It uses convolutional layers to learn spatial hierarchies of features from the input data automatically and adaptively.

\textbf{Bert} \cite{devlin2018bert}: It is a transformer-based method that has achieved state-of-the-art results on a wide range of natural language processing tasks. It uses a masked language model objective to pre-train deep bidirectional representations from unlabeled text.

\noindent \textbf{(3) Graph Based Representation Learning Models}

\textbf{TextGCN} \cite{yao2019graph}: TextGCN is a graph-based method for text classification that constructs a single, large graph over all documents in the corpus. It then applies a Graph Convolutional Network (GCN) to this graph to learn document representations.

\textbf{GAT} \cite{velivckovic2017graph}: It uses attention mechanisms to capture the importance of neighbors in the graph, which has been used for various tasks, including node and graph classification.

\textbf{TextING} \cite{zhang2020every}: TextING is a graph-based method for text classification that constructs a text information graph and applies a graph neural network to learn representations.

\textbf{DGI} \cite{velickovic2019deep}: An unsupervised graph embedding algorithm based on mutual information whose goal is to maximize the mutual information between a local representation (patch) and the corresponding graph summary representation (summary).

\textbf{GMI} \cite{peng2020graph}: GMI is a method for unsupervised learning on graphs. It uses mutual information to measure the dependency between the input and output of a graph neural network.

\textbf{TGNCL} \cite{li2023graph}: It builds a graph for each document and develops a contrastive learning regularization to learn fine-grained word representations.

\begin{table*}[h]
	
	\renewcommand\arraystretch{1.3}
	\setlength{\tabcolsep}{7.5pt}
	\caption{The test accuracy and F1 score of different methods on four datasets.}

	\begin{center}
			
		{
			\begin{tabular}{ccccccccccccc}
				\hline
				\multirow{2}{*}{\textbf{Method}}&&\multicolumn{2}{c}{\textbf{AG News}}&&\multicolumn{2}{c}{\textbf{20NG}}&&\multicolumn{2}{c}{\textbf{SougouNews}}&&\multicolumn{2}{c}{\textbf{THUCNews}} \\
				\cline{3-4} \cline{6-7} \cline{9-10} \cline{12-13}
				~ && \textbf{\textit{P}}& \textbf{\textit{F1}} && \textbf{\textit{P}}& \textbf{\textit{F1}} && \textbf{\textit{P}}& \textbf{\textit{F1}} && \textbf{\textit{P}}& \textbf{\textit{F1}} \\
				\hline
                \textbf{TF-IDF+LR}  &  & 85.92 & 85.21 & & 83.19 & 82.56 & & 86.12 & 85.35 & & 89.97 & 88.16 \\

                \textbf{FastText}  &  & 87.17 & 87.05 & & 79.38 & 78.47 & & 82.98 & 81.73 & & 86.46 & 84.08 \\

				\textbf{CNN}  &  & 88.21 & 86.43 & & 76.96 & 75.83 & & 93.64 & 93.25 & & 92.73 & 92.4 \\
			
				\textbf{Bert}  &  & 91.34 & 90.61 & & 86.54 & 86.13 & & 97.22 & 96.94 & & 96.77 & 96.41 \\
		
				\textbf{TextGCN} &  & 89.61 & 88.92 & & 85.27 & 84.49 & & 97.34 & 97.02 & & 96.82 &  96.6 \\

				\textbf{TextING}  &  & 90.52 & 89.75 & & 85.74 & 84.93 & & 96.97 & 96.48 & & 97.32 & 96.89 \\
				
				\textbf{GAT}  &  & \textbf{92.23} & \textbf{91.67} & & 86.19 & 85.73 & & 97.84 & 97.33 & & 97.54 & 97.28 \\

				\textbf{DGI}  &  & 91.4 & 90.76 & & 85.96 & 85.15 & & 96.43 & 96.01 & & 95.87 & 95.54 \\
				\textbf{GMI}\  &  & 90.95 & 89.82 & & 86.23 & 85.46 & & 96.62 & 96.16 & & 96.25 & 95.88 \\

				\textbf{TGNCL}\  &  & 89.47 & 86.91 & & 85.92 & 85.13 & & 96.37 & 85.18 & & 94.1 & 93.27 \\
				\textbf{SE-GCL} &  & 91.56 & 90.92 & & \textbf{86.76} & \textbf{85.92} & & \textbf{98.03} & \textbf{97.58} & & \textbf{97.79} & \textbf{97.32} \\
				\hline
			\end{tabular}
		}
		\label{tab1}
	\end{center}
\end{table*}

These comparative methods were chosen due to their relevance and performance in text representation learning tasks. By comparing the SE-GCL method against these methods, we aim to provide a comprehensive evaluation of its effects. It is of significance to note that both TextGCN and TextING possess their own distinctive composition methodologies, and we shall employ the techniques delineated in their original papers for text classification. Furthermore, given that GAT, DGI, and GMI are purely graph neural network algorithms, we will process them based on the intra-relation graph we have constructed to procure the corresponding text representation. Remarkably, it can also be viewed as an ablation experiment designed to validate the efficacy of our proposed contrastive learning framework.

\subsection{Experimental Setup}
All experiments were conducted using the PyTorch framework, a popular open-source machine-learning library for Python. The experiments were run on a computer equipped with an i7-9700kf CPU and an RTX2080s GPU, ensuring sufficient computational resources for the tasks.

During the model training phase, we trained all models until the loss value converged to ensure optimal results. To account for variability and randomness in the training process, each experiment was repeated ten times using different random seeds. The best precision and F1 scores obtained from each experiment were then averaged to provide the final result.

For large-scale datasets, we adopted a mini-batch strategy to address potential out-of-memory issues. This strategy involves dividing the dataset into smaller subsets or 'mini-batches' that are processed independently. This approach not only helps to manage memory usage but also can lead to faster and more stable convergence of the model.

In the SE-GCL method, we set the output dimension of each neuron in the hidden layer to 128. The learning rate, a critical parameter that determines the step size at each iteration while moving towards a minimum of a loss function, was set in the range of [0.005, 0.01]. The weight decay, a regularization technique that prevents the weights from growing too large, was set within the range of [0, 0.0001] for all datasets. The regularization factor was set to $n = 1e-6$, and the dropout rate was set to 0.4.

For all datasets, we allocated 70\% of the data for training and the remaining 30\% for testing. This split ensures that the models have sufficient data to learn from while providing an independent subset of data to evaluate their performance.

To ensure a fair comparison, we used the parameter settings from the original models for all comparative analyses. This ensures that each model is evaluated under its optimal conditions, providing a reliable basis for comparison.

\subsection{Experimental Results and Analysis}
Table \ref{tab1} presents the comparative evaluation of the proposed SE-GCL method against several state-of-the-art methods on four datasets: AG News, 20NG, SougouNews, and THUCNews. Precision (P) and F1 score (F1) are the evaluation metrics. It is important to note that the methods compared span both supervised learning techniques (such as CNN, Bert, TextGCN, GAT, and TextING) and unsupervised learning techniques (such as DGI, GMI, and our proposed SE-GCL). Despite this, the SE-GCL method consistently performs well across all datasets, even outperforming most supervised learning methods that leverage label information.

\begin{table*}	
	\renewcommand\arraystretch{1.3}
	\setlength{\tabcolsep}{5pt}
	\caption{Ablation experiment without event embedding, structure embedding and upper bound.}
	\begin{center}
		
		{
			\begin{tabular}{ccccccccccccc}
				\hline
				\multirow{2}{*}{\textbf{Method}}&&\multicolumn{2}{c}{\textbf{AG News}}&&\multicolumn{2}{c}{\textbf{20NG}}&&\multicolumn{2}{c}{\textbf{SougouNews}}&&\multicolumn{2}{c}{\textbf{THUCNews}} \\
				\cline{3-4} \cline{6-7} \cline{9-10} \cline{12-13}
				~ && \textbf{\textit{P}}& \textbf{\textit{F1}} && \textbf{\textit{P}}& \textbf{\textit{F1}} && \textbf{\textit{P}}& \textbf{\textit{F1}} && \textbf{\textit{P}}& \textbf{\textit{F1}} \\
				\hline
				
				\textbf{without structure}   &  & 89.72 & 88.28 & & 85.87 & 85.06 & & 96.63 & 96.05 & & 96.35 & 95.81 \\
				
				\textbf{without event}  &  & 91.24 & 89.73 & & 86.26 & 85.15 & & 97.37 & 96.56 & & 96.97 & 96.19 \\
				\textbf{without upper bound}	  &  & 91.18 & 89.71 & & 86.53 & 85.26 & & 97.49 & 96.84 & & 97.28 & 96.61 \\		
				
				\textbf{SE-GCL}            &  & \textbf{91.56} & \textbf{90.92} & & \textbf{86.76} & \textbf{85.92} & & \textbf{98.03} & \textbf{97.58} & & \textbf{97.79} & \textbf{97.32} \\
				\hline
			\end{tabular}
		}
		\label{tab2}
	\end{center}
\end{table*}

From the results, it can be observed that the SE-GCL method consistently performs well across all datasets. Specifically, on the 20NG dataset, SE-GCL achieves the highest precision of 86.76\% and the highest F1 score of 85.92\%. Similarly, on the SougouNews dataset, SE-GCL outperforms all other methods, achieving a precision of 98.03\% and an F1 score of 97.58\%. On the THUCNews dataset, SE-GCL again leads with a precision of 97.79\% and an F1 score of 97.32\%. These demonstrate the effectiveness of our novel approach in capturing the semantic and structural complexity inherent within the texts. Although on the AG News dataset, SE-GCL does not achieve the highest precision (which is achieved by GAT at 92.23\%), it still delivers a competitive performance with a precision of 91.56\% and an F1 score of 90.92\%. This could be attributed to the fact that AG News may contain less of the text-event information that our method is designed to capture.

Our experiments reveal that TF-IDF+LR exhibits excellent performance across all tested datasets, particularly on the 20NG, where its performance is on par with other strong baseline models. On the other hand, fastText shows promising results on the AG News dataset but suffers from performance drops on the Chinese News dataset. We speculate that this may be because fastText learns some less discriminative representations when processing longer texts, thus affecting its performance. For traditional deep learning models, CNN achieves promising results compared to the baseline on the AG News dataset. However, the performance on other datasets is obviously not as good as the results of other baselines, which shows that CNN can model short-range semantics and continuous semantics, but it does not have advantages in long texts. Bert also treats text as a sequence of words and performs significantly better than CNN on four datasets, achieving competitive results even against strong baseline methods. This shows that Bert can capture the long-range semantic relation of sequences through the self-attention mechanism. We observe that graph-based models achieve more competitive results, indicating that graph models are beneficial for text processing. TextGCN performs worse than TextING on AG News, 20NG, and THUCNews. This may be related to the inability of the corpus-level graph to explore the semantic structure information within the text. It is worth noting that GAT achieves the best results on AG News, which can benefit from the attention mechanism’s ability to capture more important semantic information. For self-supervised graph contrastive learning methods, we note that the TGNCL model achieves performance levels comparable to some semi-supervised methods. However, its performance failed to surpass the SE-GCL model on all test datasets. This observation implies that the complex data augmentation techniques adopted in TGNCL may have resulted in a certain degree of semantic information loss, thereby affecting the overall performance of the model. Furthermore, SE-GCL outperforms the self-supervised methods DGI and GMI on all datasets. The most significant improvement is observed on the THUCNews dataset, where SE-GCL achieves a 1.92\% improvement in accuracy compared to DGI.

In conclusion, the experimental results provide strong empirical evidence supporting the effectiveness and robustness of the SE-GCL method for text representation learning. Its novel approach, which includes the construction of an intra-relation graph, event skeleton extraction, and the event-based contrastive framework, leads to improved performance across various datasets, outperforming both supervised and unsupervised methods.

\subsection{Ablation Experiments}
In this ablation experiment, we investigate the impact of removing specific components of the SE-GCL method, specifically the structure embedding, event embedding (i.e., event skeleton information), and the upper bound loss.

From the results in Table \ref{tab2}, we can observe that each component of the SE-GCL method contributes to its performance. The performance drops when any of the components is removed, indicating their importance in the method. When the structure embedding is removed, the precision and F1 score on the AG News dataset drop from 91.56\% and 90.92\% to 89.72\% and 88.23\%, respectively. Similar drops in performance are observed on the other datasets. This indicates that the structure information contributes significantly to the effectiveness of the SE-GCL method.

Removing the event embedding also leads to a decrease in performance, but the impact is less pronounced than removing the structure embedding. For instance, on the THUCNews dataset, the precision and F1 score drop from 97.79\% and 97.32\% to 96.97\% and 96.19\%, respectively. This suggests that the event skeleton information, while important, is less critical than the structure information. The upper bound loss also plays a role in the performance, but its removal has a less pronounced impact on the results. This suggests that while the upper bound loss contributes to the performance of the SE-GCL method, it is not as critical as the structure and event embeddings.

It merits attention that the impact of eliminating event embedding on the AG News dataset surpasses that of removing the upper bound, a finding that stands in contrast to the results observed in the other three datasets. Upon scrutinizing the outcomes of the preceding comparative experiments, we conjecture that this discrepancy can be attributed to the brevity of the text length in AG News and the consequent scarcity of event information. Consequently, the removal of event embedding does not significantly influence the results.

These findings highlight the importance of structure embedding, event embedding, and upper bound loss in achieving high performance in text representation learning with the SE-GCL method. They also underscore the effectiveness of the SE-GCL method, which outperforms all ablated versions on all datasets.

\subsection{Parameter Analysis}
In our experiments, we conducted a detailed analysis of the hyperparameters, including $\eta$ and $\theta$ in eq. (9) as well as $W_e$ and $W_s$ in eq. (10).

\begin{figure}
	\vspace{-3pt}
	\centering
	\includegraphics[width=0.49 \textwidth]{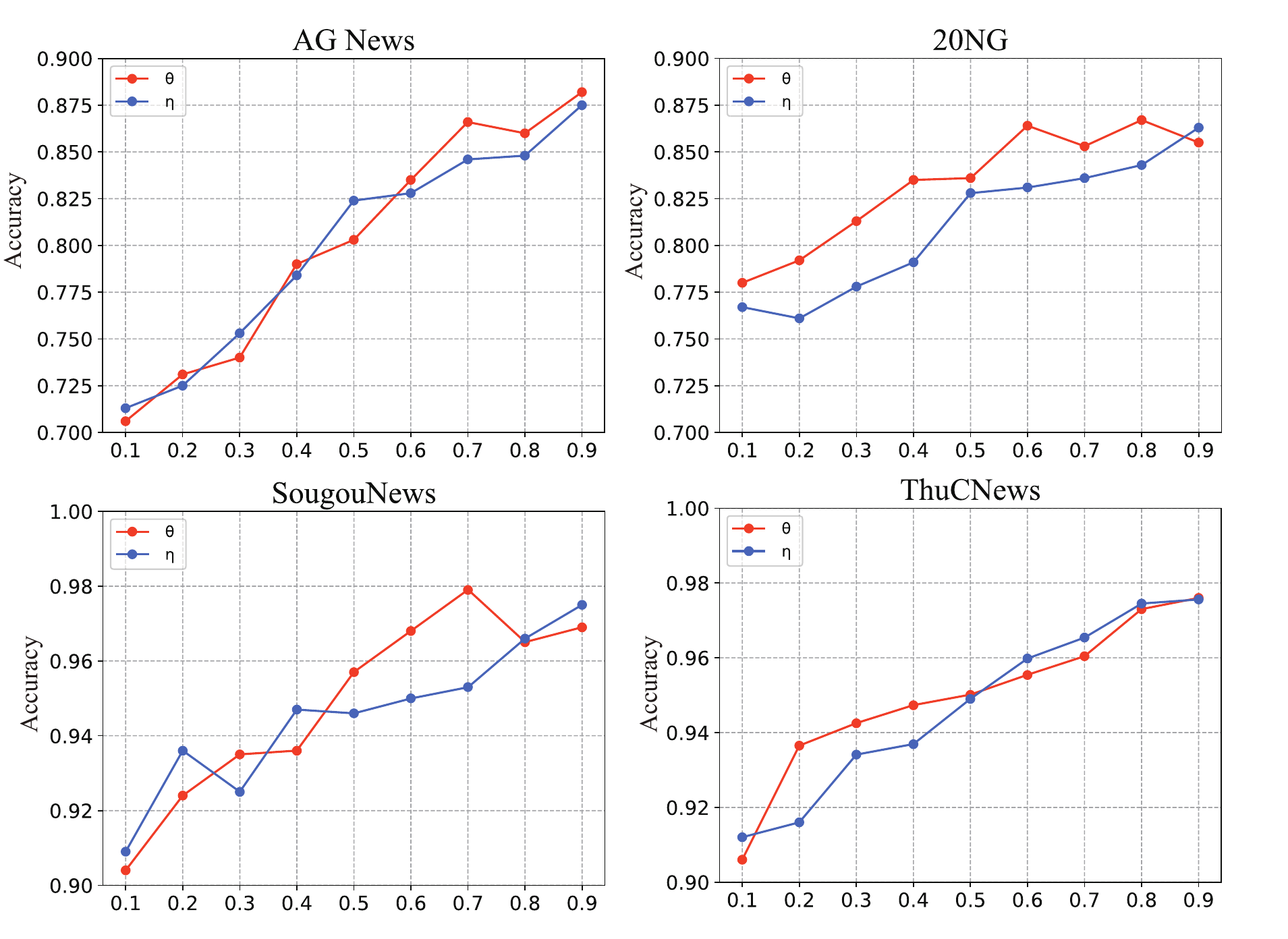}
	\caption{Experiment results in different settings ($\theta$ and $\eta$).}
	\label{Fig4}
\end{figure}

For $\eta$ and $\theta$, we set their values in the range [0.1, 0.9]. While adjusting one parameter, the other parameter was held constant at its optimal value of 0.9. The experimental results, as shown in Fig. \ref{Fig4}, indicate that the performance is poor when their values are small, and the performance improves as the value increases until it reaches a relatively stable state. This can be attributed to the fact that when their values are small, the difference between positive and negative pairs is too small, resulting in insufficient discrimination.

\begin{figure}
	\centering
	\includegraphics[width=0.49 \textwidth]{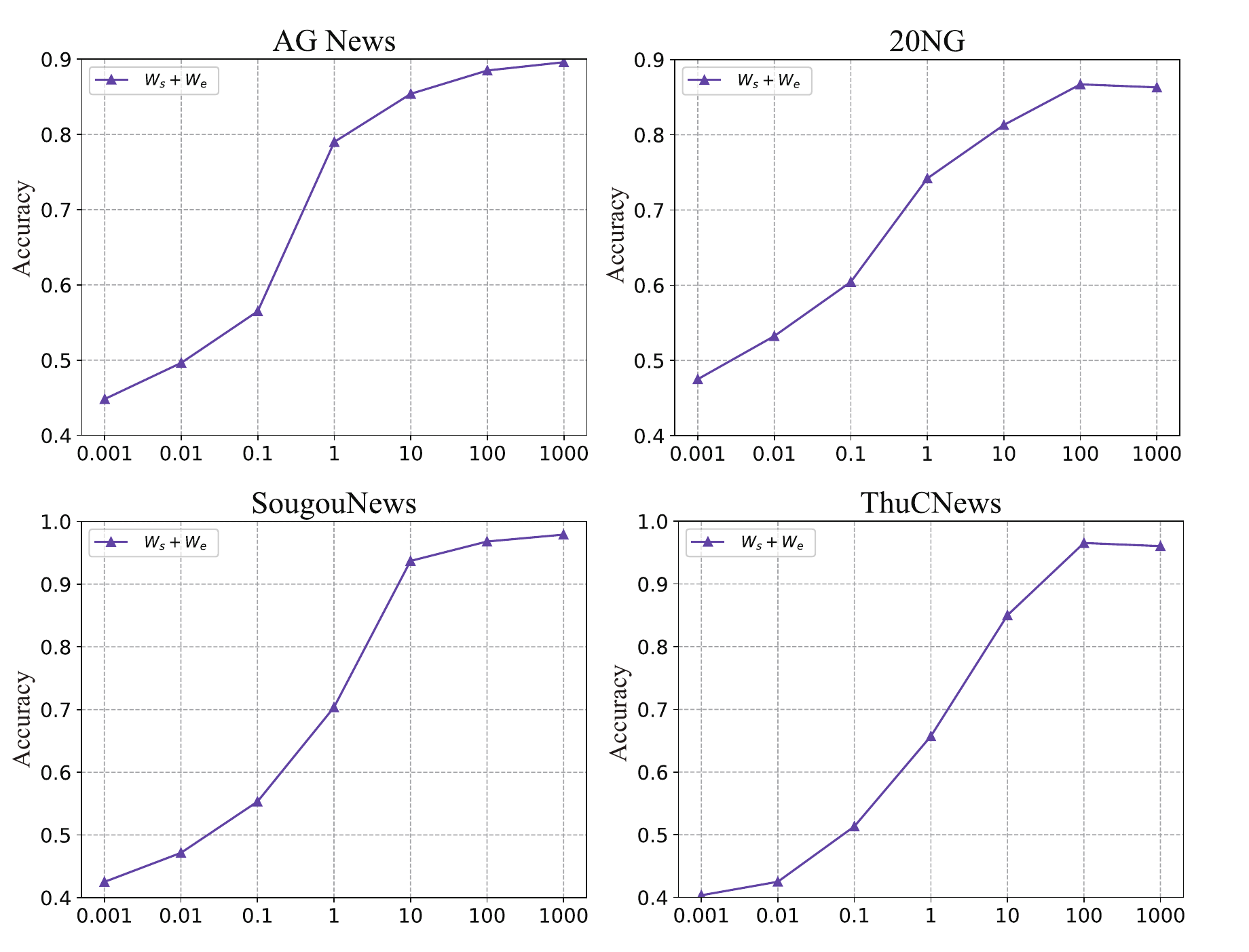}
	\caption{Experiment results in different settings ($W_e$ and $W_s$)}
	\label{Fig5}
\end{figure}

Similarly, parameters $W_e$ and $W_s$ were adjusted between [0.001, 1000], with one parameter held constant at its optimal value during the adjustment of the other. As depicted in Fig. \ref{Fig5}, the experimental results are poor when their values are small. This suggests that $\zeta_e$ and $\zeta_s$ are important for the performance of our method.

These findings underscore the importance of carefully selecting the hyperparameters in our SE-GCL method. They also highlight the effectiveness of our method, which achieves high performance across a range of hyperparameter settings.

\subsection{Efficiency Analysis}
In our experiments, we also conducted an efficiency analysis to compare the time consumption of our SE-GCL method with other methods, including CNN, DGI, GAT, GMI, and BERT.

\begin{figure}
	\centering
	\includegraphics[width=0.49 \textwidth]{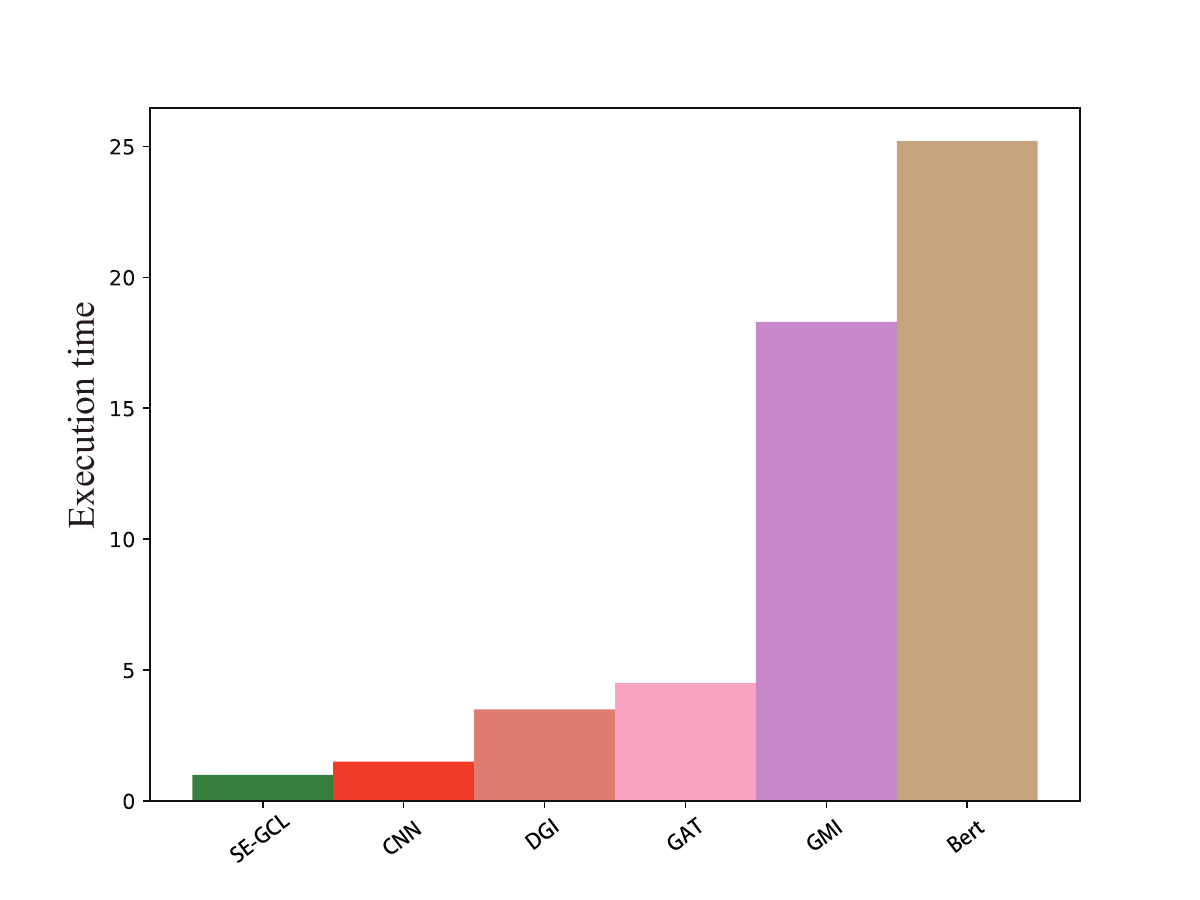}
	\caption{Comparison of execution time of different methods. The execution time of our SE-GCL method is 1.}
	\label{Fig6}
\end{figure}

The time consumption of each method is presented in Fig. \ref{Fig6}. For the purpose of comparison, we set the time consumption of SE-GCL as 1. The time consumption of the other methods is as follows: CNN (1.5), DGI (3.5), GAT (4.7), GMI (18.6), and BERT (24.8).

From the results, it is evident that our SE-GCL method is more efficient than all the other methods. Specifically, our proposed unsupervised learning SE-GCL demonstrates algorithmic advantages compared to semi-supervised methods.

Furthermore, the intra-relation graph we pass into the model is simpler than the graph neural network approach, which contributes to the efficiency of our method.

In terms of the unsupervised contrastive learning method, we replace the discriminator by setting an upper bound loss, obtain event embeddings through MLP, structural embeddings through GCN, and negative embeddings through shuffling anchor embeddings. These strategies effectively reduce the consumption of the algorithm, further enhancing the efficiency of our SE-GCL method.

These findings highlight the efficiency of our SE-GCL method, which achieves high performance with less time consumption compared to other methods.

\section{Conclusion}
In this paper, we introduced the Event-based Graph Contrastive Text Representation Learning (SE-GCL) method. SE-GCL effectively captures both the semantic and structural intricacies inherent in natural language through a systematic pipeline comprising four major steps: intra-relation graph construction, event skeleton extraction, embedding generation in a contrastive framework, and the employment of multiple loss functions. First, grasp the core purpose of the text semantically and structurally by identifying event elements in the text and constructing intra-relation graphs for each text. Then, more representative textual semantic information is captured by extracting the event skeleton of the intra-relation graph. Besides, we explore the complementarity between event information and structural information through positive embeddings constructed from different perspectives. Among them, we have greatly simplified the embedding generation method, improving efficiency while ensuring the effect. Finally, multiple loss functions are used to expand the inter-class differences in embeddings while reducing the intra-class differences. Our experimental results on four real-world datasets show that SE-GCL outperforms several state-of-the-art methods in terms of precision and F1 score. Furthermore, our ablation study highlights the importance of each component, while efficiency analysis shows that SE-GCL is more time-saving than other methods. Considering that the SE-GCL model incorporates the concept of events, it is particularly well-suited for processing medium to long texts containing rich event elements. However, the nature of such texts often comes with complexity, a characteristic common in multi-label classification datasets. Currently, our model architecture is not directly optimized for such multi-label scenarios, which limits its applicability to some extent. Given the significant research value of multi-label classification tasks, we plan to expand the SE-GCL model in future work to accommodate the needs of multi-label classification. All in all, these findings underscore the effectiveness and efficiency of SEGCL in text representation learning, paving the way for its application in various natural language processing tasks.

\section{Acknowledgements}
This work is supported by National Natural Science Foundation of China (Grant No. 69189338), Excellent Young Scholars of Hunan Province of China (Grant No. 22B0275), and Changsha Natural Science Foundation (Grant No. kq2202294).
\backmatter

\section*{Statements and Declarations}
\textbf{}

\textbf{Conflict of interests} The authors declare that they have no conflict of interest.

\textbf{Data Availability} The datasets used during the current study are available on the Kaggle platform link \href{https://www.kaggle.com}{https://www.kaggle.com} and online repositories  \href{http://qwone.com}{http://qwone.com} and \href{https://github.com}{https://github.com/}.

\textbf{Authors contributions} \textbf{Tao Meng}: Investigation, Conception and design of study, Acquisition of data, Software, Writing the original manuscript. \textbf{Wei Ai}: Funding acquisition, Software, Validation, Reviewing and Editing. \textbf{Jianbin Li}: Funding acquisition, Resources, Reviewing and Editing. \textbf{Ze Wang}: Methodology, Analysis and interpretation of Results, Writing-Reviewing and Editing, Supervision. \textbf{Keqin Li}: Reviewing and Editing.

\bibliography{sn-bibliography}

\end{document}